\documentclass[runningheads]{llncs}
\usepackage{graphicx}
\usepackage{amsmath,amssymb} 
\usepackage{color}

\usepackage{xcolor}
\usepackage{booktabs} 
\usepackage{times}
\usepackage{epsfig}
\usepackage{subfigure}
\usepackage{siunitx}
\usepackage{textcomp}
\usepackage[utf8]{inputenc}
\usepackage[english]{babel}
\usepackage{url}
\usepackage{colortbl}
\usepackage[caption=false]{subfig}

\begin{document}
\title{Learning from \#Barcelona Instagram data what Locals and Tourists post about its Neighbourhoods} 

\titlerunning{Learning from \#Barcelona Instagram data}
%
\author{Raul Gomez\inst{1,2}\orcidID{0000-0003-4460-3500} \and
Lluis Gomez\inst{2}\orcidID{0000-0003-1408-9803} \and
Jaume Gibert\inst{1}\orcidID{0000-0002-9723-3913} \and
Dimosthenis Karatzas\inst{2}\orcidID{0000-0001-8762-4454}}
%
\authorrunning{R. Gomez, L. Gomez, J. Gibert, D. Karatzas}
%

\institute{
Eurecat, Centre Tecnològic de Catalunya, Unitat de Tecnologies Audiovisuals, Barcelona, Spain\\
\email{\{raul.gomez,jaume.gibert\}@eurecat.org}\and
Computer Vision Center, Universitat Autònoma de Barcelona, Barcelona, Spain\\
\email{\{lgomez,dimos\}@cvc.uab.es}} 

\maketitle              
\begin{abstract}
Massive tourism is becoming a big problem for some cities, such as Barcelona, due to its concentration in some neighborhoods.
In this work we gather Instagram data related to Barcelona consisting on images-captions pairs and, using the text as a supervisory signal, we learn relations between images, words and neighborhoods. Our goal is to learn which visual elements appear in photos when people is posting about each neighborhood.
We perform a language separate treatment of the data and show that it can be extrapolated to a tourists and locals separate analysis, and that tourism is reflected in Social Media at a neighborhood level.  
The presented pipeline allows analyzing the differences between the images that tourists and locals associate to the different neighborhoods.


The proposed method, which can be extended to other cities or subjects, proves that Instagram data can be used to train multi-modal (image and text) machine learning models
that are useful to analyze publications about a city at a neighborhood level. We publish the collected dataset, \emph{InstaBarcelona} and the code used in the analysis. 

\end{abstract}
\section{Introduction}






Instagram is an image based social network where people tend to post high quality personal pictures accompanied by a caption. Captions are diverse, but they usually describe the photo content, the place where the photo was taken or the feelings the photo brings in. The objective of adding this text, which usually contains hashtags, is that other Instagram users can find the photo using one of the words and follow the author if they like what they post. The number of images updated to Instagram is huge: If we search for images accompanied by the word "Barcelona" we find more than 1 million.

\begin{figure}
\begin{minipage}[c]{0.48\linewidth}
  \includegraphics[width=1\linewidth]{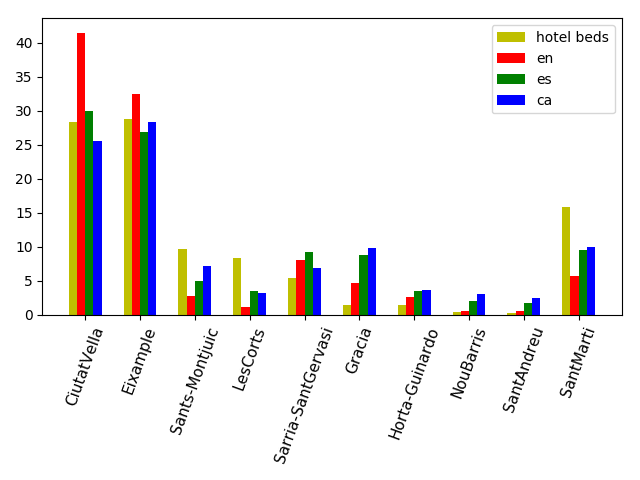}
   \caption{\% of mentions per district respect to the total districts mentions in each language. In yellow, the \% of hotel beds per district.}
   \label{fig:districts}
\end{minipage}
\hfill
\begin{minipage}[c]{0.48\linewidth}
  \includegraphics[width=1\linewidth]{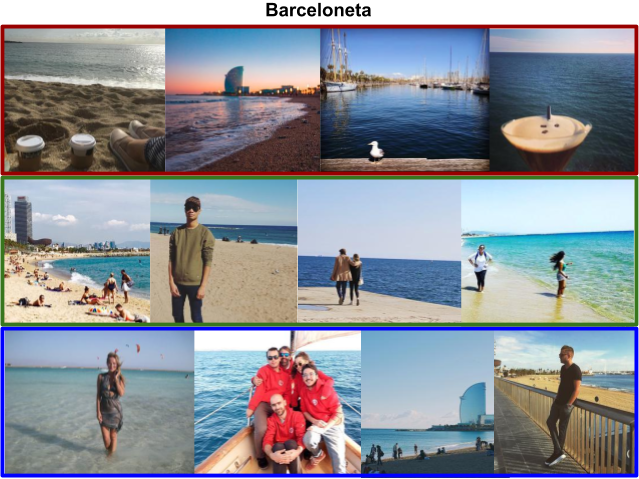}
   \caption{Img2NeighCtx image by neighborhood retrieval results for \emph{La Barceloneta} in English (top), Spanish (middle) and Catalan (bottom).}
   \label{fig:txt2img_barceloneta}
\end{minipage}%
\end{figure}

This work shows how Instagram data can be exploited to obtain information about a 
city that has very interesting social and commercial applications. Specifically, we analyze images and captions related to Barcelona. Barcelona is a very touristic city which revives around $10$ million tourists every year. That causes conflicts between tourists and locals and between the tourism industry and other local organizations, conflicts that are highly concentrated on neighborhoods with requested tourist attractions. Measuring the tourism overcrowding per neighborhood is not easy, since some areas receive high touristic interest but they don't have hotels or tourism installations. This work proposes a method to do that by exploiting Instagram data. 

We perform a multi-modal, language separate analysis using the text of the captions and its associated images, designing a pipeline that learns relations between words, images and neighborhoods in a self-supervised way. We focus on a per-neighborhood analysis, and analyze how the differences of tourism activity between Barcelona districts and neighborhoods are reflected on Instagram. 
Notice that, despite in this work we apply the proposed pipeline to Barcelona, it is extensible to any other city with enough Social Media activity to collect the required data. 
The proposed method works as follows:
\begin{enumerate}
\item We split the data depending on whether it contains captions in a local language, Spanish and Catalan, or English, which we consider to be locals vs tourists publications (Section \ref{sec:3}).
\item We count the mentions of the different districts and neighborhoods and the most used words in each data split. The results confirm that the language-separate treatment can be extrapolated to a locals vs tourists analysis (Sections \ref{sec:41}, \ref{sec:42}).
\item We train a semantic word embedding model, Word2Vec \cite{Mikolov2013}, for each language and show the words that locals and tourists associate with different neighborhood names (Sections \ref{sec:43}, \ref{sec:44}).
\item Using the semantic word embeddings as a supervisory signal, we train a CNN than learns relations between images and neighborhoods. (Section \ref{sec:51})
\item Using the trained models in a retrieval approach with unseen data we show, for each language, the most related images to different neighborhoods. Results show interesting differences between the visual elements that locals and tourists associate to each neighborhood. (Section \ref{sec:52})
\end{enumerate}

The contributions of this work are as follows: First, we show how state of the art multi-modal text and images techniques can be applied to learn from Instagram data. Second, we show how Instagram data related to a city can be used to do a per-neighborhood analysis obtaining very useful social and commercial information. Specifically, we propose a method to analyze tourism activity at a neighborhood level using only images and text. Third, we provide a new dataset, \emph{InstaBarcelona}, formed by Instagram images related to Barcelona and its captions, and the code and models used to perform the subsequent experiments.

\section{Related Work}
Deep learning advances and the availability of "free" Web and Social Media data have motivated the research of pipelines that can learn from images and associated text in a self-supervised way. In order to do that, state of the art algorithms vectorize text using word embedding methods, such as Word2Vec \cite{Mikolov2013} and GloVe \cite{Pennington}, topic models \cite{Blei2003} or LSTM encodings. Then features are extracted from images using a CNN, and a model is trained to learn relations from those representations.   
This pipeline was originally proposed by A. Frome \textit{\textit{et al.}} with DeVISE \cite{Frome2013} which, instead of learning to predict ImageNet classes, it learns to infer the Word2Vec \cite{Mikolov2013} representations of their labels.
Later, Gordo \& Larlus \cite{DianeLarlus2017} used captions associated to images to learn a common embedding space for images and text through which they perform semantic image retrieval.
Learning from Web data, Gomez \textit{\textit{et al.}} \cite{Gomez2017} use LDA \cite{Blei2003} to extract topic probabilities from a bunch of Wikipedia articles and train a CNN to embed its associated images in the same topic space.
In a more specific application, Salvador \textit{\textit{et al.}} \cite{Salvador} collected data from cooking websites and proposed a joint embedding of food images and its recipes to identify ingredients, using Word2Vec \cite{Mikolov2013} and LSTM representations to encode ingredient names and cooking instructions and a CNN to extract visual features from the associated images.

Social Media data has already been exploited in city analysis. 
In \cite{Engineering2017}, Instagram uses publications texts and geolocations to find neighborhoods with similar activity across different United States cities. To do that they look for words shared between cities but not between neighborhoods of a city, train a topic model \cite{Blei2003} and find neighborhoods that share topics. 
S. Chang \cite{Chang} exploits similar data to analyze popular hashtags in different locations and show cultural differences between different cities and neighborhoods. 
J. Boy \textit{et al.} \cite{Boy} analyze Instagram Amsterdam activity to study how the different neighborhoods, events and cultures of the city are represented in Instagram. 
Y. Kuo \textit{et al.} \cite{Kuo} mine data from different sources (Instagram, Twitter, TripAdvisor, etc.) and modalities (images, text and geolocations) related to New York City to analyze citizens behavior in differents aspects such as trends, food or transportation. 
V. Singh \textit{et al.} \cite{Singh2017} detect raze, age and gender of people in New York City Instagram images to analyze social diversity of different neighborhoods and compare it to census-based metrics.
With a more similar objective as ours, Garcia-Palomares \textit{et al.} \cite{Garcia-Palomares2015} use geolocated Social Media photographs on Panoramio to identify the main tourist attractions in eight major European cities. They compare tourist and locals activity and study the distribution of it over the city. However, our work differs in the data, since they use geolocations in their analysis instead of images and text.

To our knowledge, this is the first work exploiting multi-modal image-text Social Media data to do a per neighborhood analysis of a city comparing tourism vs local activity.

\section{Dataset: InstaBarcelona} \label{sec:3}
\begin{figure}
\begin{minipage}[c]{0.48\linewidth}
\centering
\includegraphics[width=0.7\linewidth]{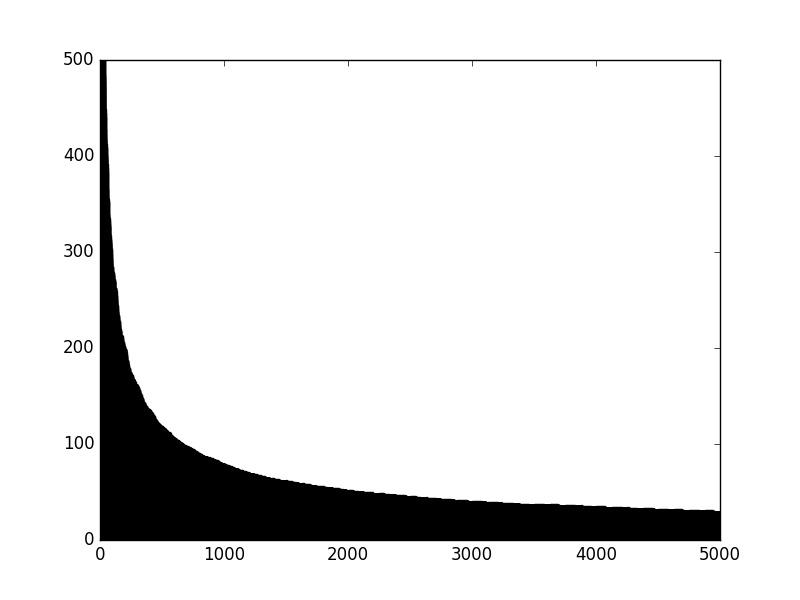}
\caption{Number of publications of top 5000 users.}
\label{fig:users}
\end{minipage}
\hfill
\begin{minipage}[c]{0.48\linewidth}
\centering
\includegraphics[width=0.7\linewidth]{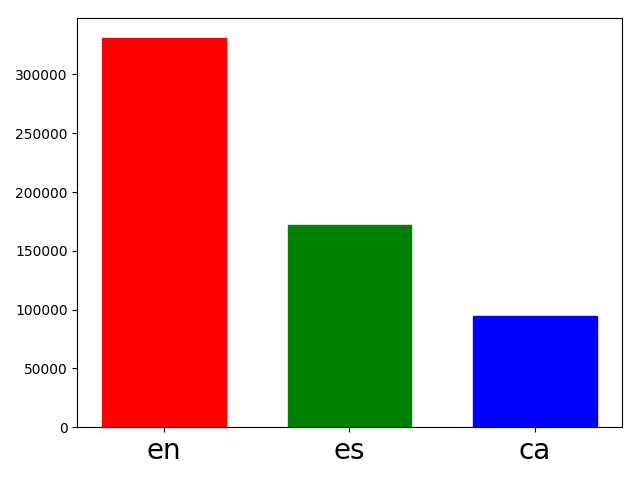}
\caption{Number of images per language.}
\label{fig:language_distribution}
\end{minipage}%
\end{figure}

To perform the presented analysis we gathered a dataset of Instagram images related to Barcelona uploaded between September and December of 2017.  That means images with a caption where the word "Barcelona" appears. We collected around $1.3$ million images. In order to discard spam and other undesirable images, we performed several dataset cleanings:
\textit{Users with many publications} tend to be spam or commercial accounts. We found $335,880$ different users, where the user with more publications has $4,357$. Figure \ref{fig:users} shows the number of publications of the top $5,000$ users. We blacklisted the users having more than $50$ publications and discarded all their content. The number of blacklisted accounts was $2,123$.
\textit{Images with short captions} are not desirable since they usually do not provide enough information of the image to learn from. We discarded all the images accompanied with captions shorter than $3$ words. 
\textit{Repeated images} tend to be spam and should be discarded. 
\textit{Images containing other city names in their captions} were also discarded, since they tend to be spam and to not provide information related to Barcelona. 
\textit{Images with captions in other languages} than English, Spanish or Catalan were discarded since in this work we analyze publications related to those languages. To infer the language of the captions Google's language detection API \footnote{\url{https://code.google.com/archive/p/language-detection/}} was used.

The resulting dataset, \emph{InstaBarcelona}, contains $597,766$ image-captions pairs and is made publicly available for download. From those pairs $331,037$ are English publications, $171,825$ Spanish publications and $94,311$ Catalan publications (Figure \ref{fig:language_distribution}). The dataset is available on \url{https://gombru.github.io/2018/08/02/InstaBarcelona/}.

\section{Textual Analysis}

\begin{figure*}[htp]
    \centering
    \includegraphics[width=0.9\linewidth]{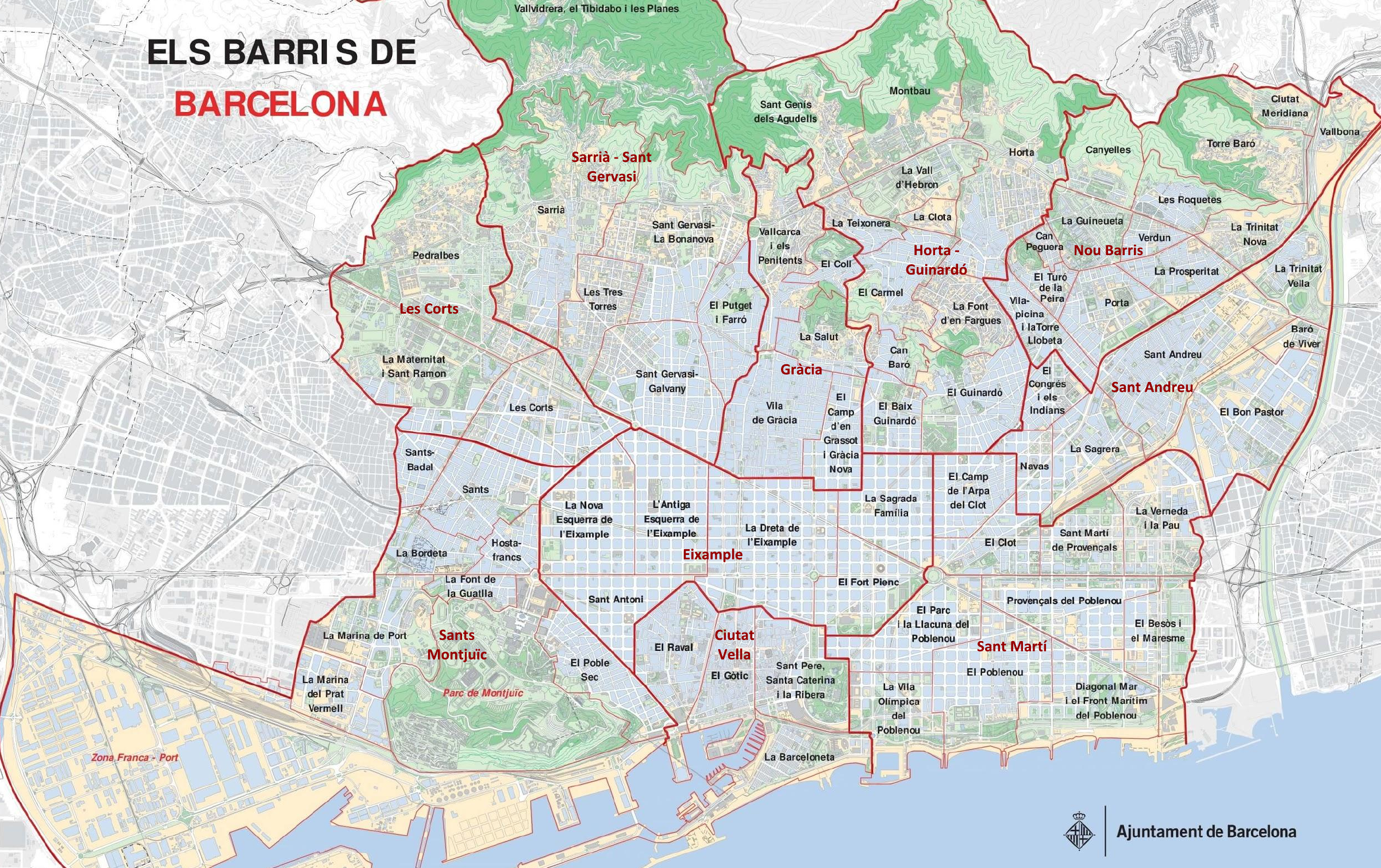}
    \caption{Barcelona map showing its districts and neighborhoods.} \label{fig:bcn}
\end{figure*}

Barcelona is divided in 10 districts, which are divided in several neighborhoods \footnote{\url{http://www.barcelona.cat/en/living-in-bcn/living-neighbourhood}}, as shown in Figure \ref{fig:bcn}. 
In this section we use districts and neighborhoods names to perform a textual analysis and show how tourism is reflected in Instagram activity related to Barcelona.

\subsection{Most Frequent Words} \label{sec:41}
\begin{figure}[htp]
    \centering
    \includegraphics[width=0.5\linewidth]{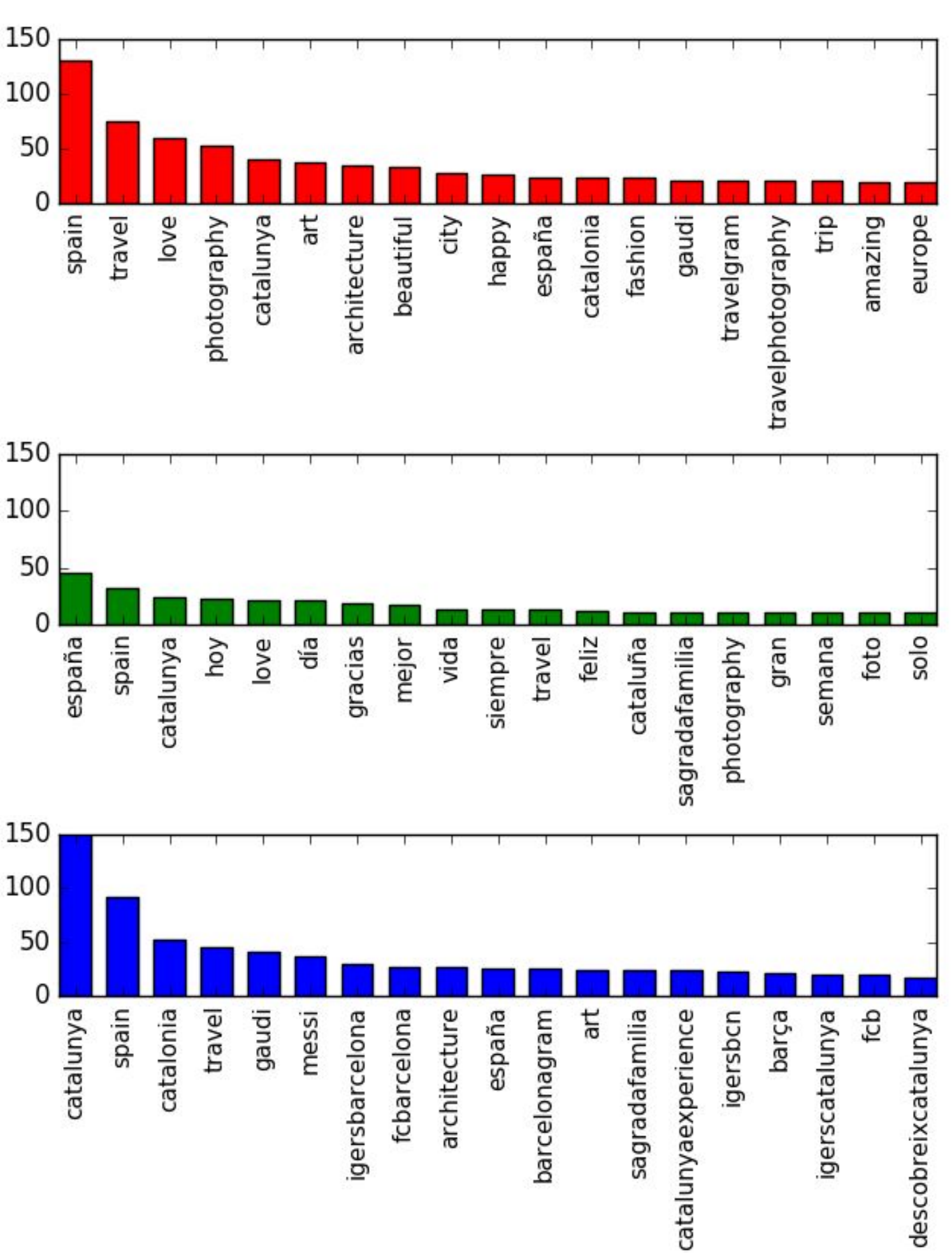}
    \caption{$\it{^{0}/_{000}}$ of most frequent words instances respect to the total words on each of the languages. Red shows English dataset results, green Spanish and blue Catalan.} \label{fig:top_words}
\end{figure}

Figure \ref{fig:top_words} shows the most used words in each language. Words without a semantic meaning (connectors, etc), "barcelona", "bcn" and words related to Nueva Barcelona del Cerro Santo city, in Venezuela, have been removed. 
While in English most of the top words are related to tourism ("travel", "photography", "art", "architecture", "trip") in the local languages other kind of terms appear in the top words ("hoy", día", "gracias", "vida", "messi, "igersbarcelona", "fcbarcelona"), which supports our assumption of considering English publications as tourists publications, and Spanish and Catalan as locals publications.


\subsection{Most Mentioned Districts} \label{sec:42}

To compute the number of mentions per district for each language, we take into account district names, its neighborhoods names, and other abbreviations and names that are often used. Figure \ref{fig:districts} shows, for each language, the \% of mentions per district respect to the total districts mentions in that language. It also shows the number of hotel beds in each district given by the city hall of Barcelona \cite{bcn_hotel_beds}. \textit{Ciutat Vella} and \textit{Eixample} are the most mentioned districts in the three languages. This makes sense since those districts concentrate the most representative and touristic Barcelona attractions, and people tend to post more on Instagram when they are traveling and to use the word Barcelona when they are uploading a Barcelona representative image. The \% of images that this most touristic districts concentrate is much bigger for English than for local languages, specially for \textit{Ciutat Vella}, Barcelona's old town, known as the most touristic district. In all the other Barcelona districts, the \% of publications is always higher for local languages than for English. The number of hotel beds is also markedly higher in \textit{Ciutat Vella} and \textit{Eixample}, which is consistent with our results. However, obtaining tourism measures for city areas is difficult. The number of hotel beds, which is the most meaningful of the tourism measures provided by Barcelona City Hall, is not necessarily correlated with tourism activity, since one district could be very visited by tourist but have few hotel beds due to, for instance, its urbanism. Over more, the City Hall does not provide this data per neighborhood. 

\begin{figure*}[h]
    \begin{subfigure}
            \centering
            \includegraphics[width=0.32\linewidth]{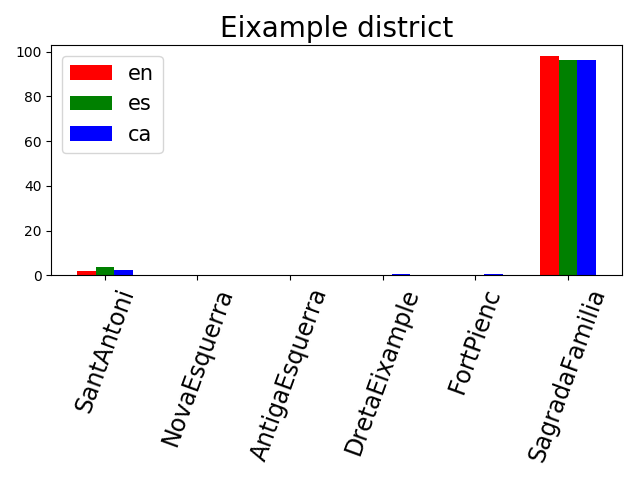}
    \end{subfigure}%
    \hfill
    \begin{subfigure}
            \centering
            \includegraphics[width=0.32\linewidth]{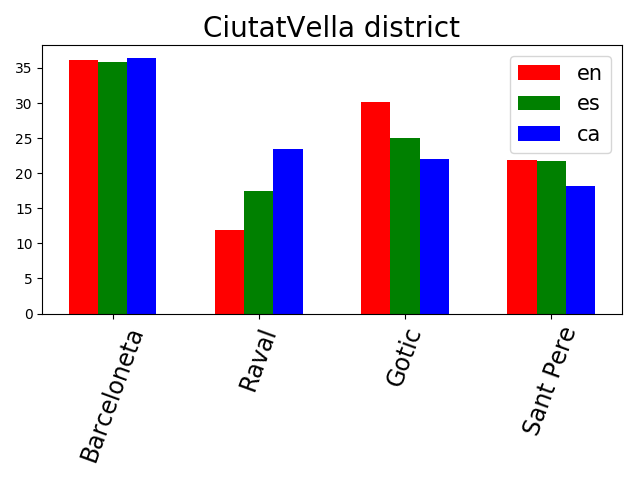}
    \end{subfigure}%
    \hfill
    \begin{subfigure}
            \centering
            \includegraphics[width=0.32\linewidth]{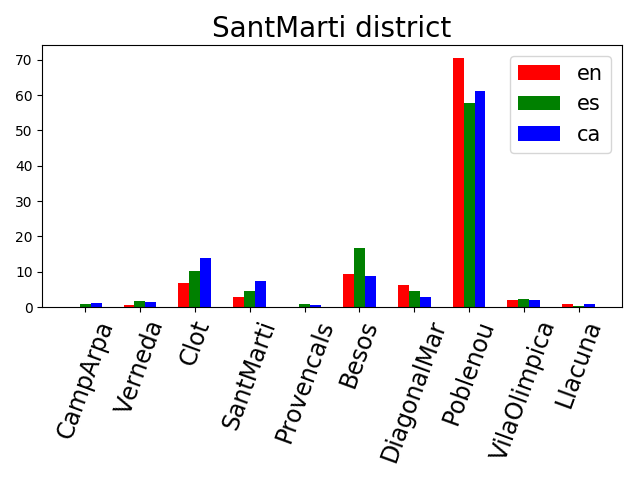}
    \end{subfigure}
    \hfill
    \begin{subfigure}
            \centering
            \includegraphics[width=0.32\linewidth]{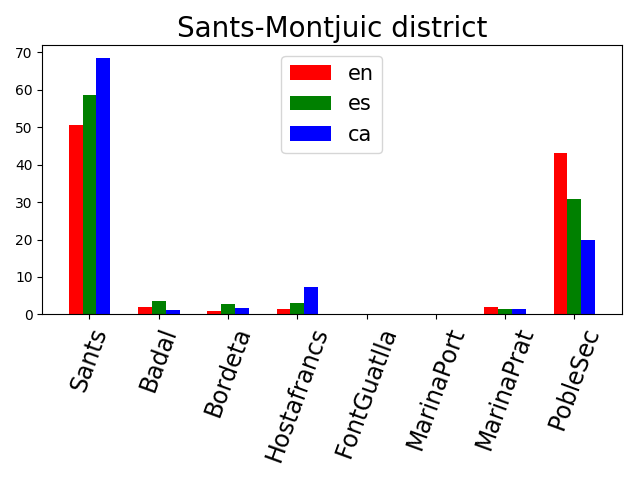}
    \end{subfigure}
    \hfill
    \begin{subfigure}
            \centering
            \includegraphics[width=0.32\linewidth]{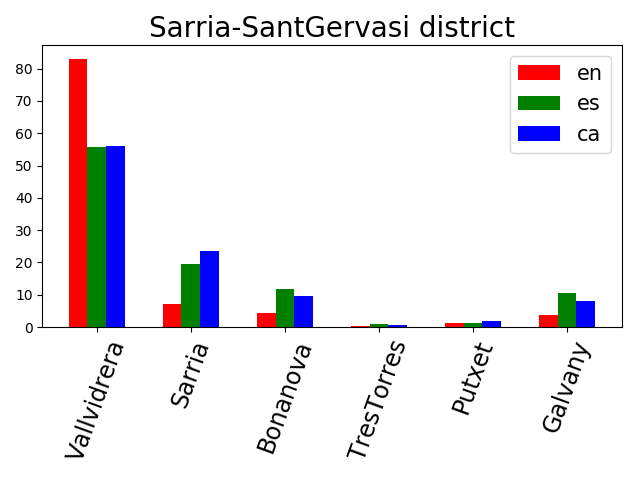}
    \end{subfigure}
    \hfill
    \begin{subfigure}
            \centering
            \includegraphics[width=0.32\linewidth]{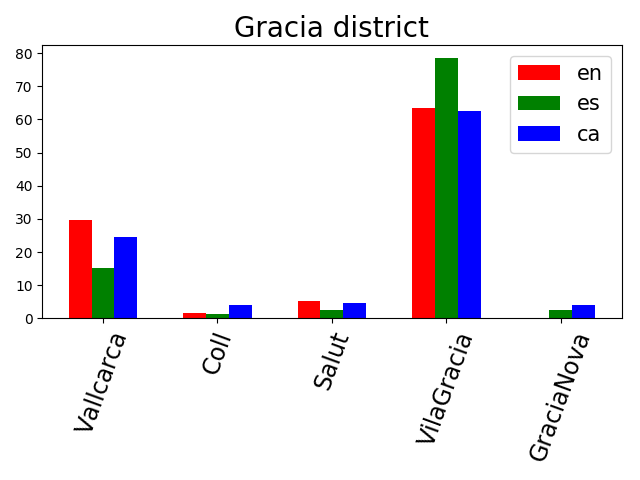}
    \end{subfigure}
    \caption{ \% of mentions per neighborhood respect to the total district neighborhoods mentions in each language.} \label{fig:neighborhoods}
\end{figure*}

Figure \ref{fig:neighborhoods} shows, for each language, the \% of mentions per neighborhood respect to the total neighborhood mentions of a district in that language. 
The \textit{Ciutat Vella} plot shows that all its neighborhoods are highly popular among all tourists and locals, being \emph{La Barceloneta}, its beach area, the most mentioned one in all languages. \emph{La Barceloneta} is a former fisher neighborhood which receives now a lot of tourism attention. 
\emph{El Gòtic}, Barcelona's old town, concentrates a markedly higher \% of publications in English than in other languages, and is in fact the neighborhood most affected by tourism in Barcelona. \emph{Sant Pere}, commonly known as \emph{El Born}, is also mentioned by tourists and locals in a similar \%. \emph{El Raval} neighborhood is a very multi-cultural area, which has traditionally been considered dangerous due to drug presence and delinquency. 
However, its geographical situation close to Barcelona's old town has transformed it lately into a more touristic area. The plot shows that \emph{El Raval} is still an area more popular among locals.

The \textit{Eixample} plot shows clearly that the only reason why this district is one of the most mentioned, specially by tourists, is the \emph{Sagrada Família}, which names both a temple and an \textit{Eixample} neighbourhood. The Sagrada Família temple is the top tourism attraction in Barcelona, and all the touristic activity in \textit{Eixample} big district is concentrated around the temple. \emph{Sant Antoni}, which is a neighborhood of increasing popularity with high probability of becoming a touristic area, is still mentioned more in local languages. The other neighborhoods are not very mentioned because they are residential areas without neighborhood identities. 

The \textit{Sant Martí} district plot, shows that \emph{El Poblenou} is the most popular neighbourhood in it, specially among english speakers. \emph{El Poblenou} is a former industrial neighbourhood which lately is getting popular due to the 22@ plan, which aims to concentrate in that area technological firms headquarters and design studies. Due to its modernization and geographical situation in the seaside, \emph{El Poblenou} is in danger to become a neighborhood with overcrowded tourism, as well as happened with \emph{La Barceloneta}. This analysis strengthens that hypothesis, showing that \emph{El Poblenou} and \emph{Diagonal Mar} are the only neighborhoods among \textit{Sant Martí} where the English \% of posts is superior to the ones of local languages.

The \textit{Sants-Montjuic} district plot shows that the most mentioned neighborhood among all languages is \emph{Sants}, which is justifiable because it's also the district name. In this district, the only neighborhood where the \% of English posts dominate over the local languages is \emph{Poble Sec}. \emph{Poble Sec} is an area besides the \textit{Ciutat Vella} district which is also overcrowded by tourism, 
as the plot indicates. Tourism influence is getting expanded across \emph{El Raval} to \emph{Poble Sec}, which is getting popular as a bar and eating area across young people and, lately, across tourists. 
The \textit{Sarrià} plot shows that its only neighborhood where the \% of English publications respect to the total is superior to the other languages is \emph{Vallvidrera}. That is because in \emph{Vallvidrera} there is the Tibidabo mountain, which attracts tourist that tend to post photos from its panoramic views.

The \textit{Gràcia} district, and specially its neighborhood \emph{Vila de Gràcia} it's a popular area, specially for young people, which is attracting many tourists lately. The plot clearly shows how \emph{Vila de Gràcia} concentrates most of the posts from the three languages.
It also shows that \emph{Vallcarca} and \emph{La Salut} neighborhoods are the ones where the \% of English posts are superior. That is because in this area the Park Güell, a big touristic claim, is located.

\subsection{Word2Vec} \label{sec:43}
Word2Vec \cite{Mikolov2013} learns vector representations from non annotated text, where words having similar semantics have similar representations. In this work we use Gensim Word2Vec implementation \footnote{\url{https://radimrehurek.com/gensim/models/word2vec.html}} and train a different model for each one of the analyzed languages: English, Spanish and Catalan. The objective is to learn the different contexts where the authors use words depending on their language. The models are trained using the CBOW mode with a dimensionality of $300$, a window of $8$ and $25$ epochs over the text corpus.

\subsection{Words Associated to Districts} \label{sec:44}
Using the Word2Vec learned models for each language, we can infer the words that users writing in English, Spanish or Catalan (tourist or locals) relate with each one of the Barcelona's neighborhoods. Next, we show the closest words in the Word2Vec space to the four \textit{Ciutat Vella} neighborhoods using the three Word2Vec models learned. Closest words of the  \textcolor{red}{English} trained Word2Vec are shown in red, of the  {\color{green!70!blue}Spanish} one in green, and of the  \textcolor{blue}{Catalan} one in blue. Spelling variants and synonyms have been removed from the results.\\ 

\begin{multicols}{2}
\scriptsize
\textit{Barceloneta}\\
\textcolor{red}{hotelw, seaside, beachlife, beachview, port, bcnbeach}\\
{\color{green!70!blue}ramblademar, torremapfre, hotelvela, paseomaritimo}\\
\textcolor{blue}{portolimpic, hotelwela, vilaolimpica, torremapfre, bogatell}\\

\textit{Gotic}\\
\textcolor{red}{cathedral, history, gargoyles, churches, architecture}\\
{\color{green!70!blue}edadmedia, laribera, carrerdelbisbe, mercadodelborn}\\
\textcolor{blue}{carrerdelbisbe, plaçadelrei, catedraldelmar, carrercomtal}\\

\textit{Born}\\
\textcolor{red}{barcelonaspots, gothicdistrict, oldtown, catedraldelmar}\\
{\color{green!70!blue}passeigdelborn, portalnou, callejuelas, rinconesmagicos}\\
\textcolor{blue}{mercatdelborn, ccm, banysorientals, cafedelborn, laribera}\\

\textit{Raval}\\
\textcolor{red}{cccb, macba, zeligbar, poblesec, elborn, grafity, gotico}\\
{\color{green!70!blue}rambladelraval, ravalcultural, fueradrogas, narcopisos}\\
\textcolor{blue}{ravalcultural, somdebarri, ravalescultura, botigadecomics}\\
\end{multicols}

This examples show the interests of the different language speakers in the query neighborhoods, 
Words related to \emph{Barceloneta} and \emph{El Gòtic} neighborhoods in the three languages are mostly tourist attractions. However, we can appreciate differences between languages. For instance, when mentioning \emph{El Gòtic}, Spanish and Catalan speakers use along names of its streets and squares, while English speakers use more general words.
Tourist publications mentioning \emph{El Born} relate this district to Barcelona's old town, while locals publications mention its promenade, its market or its culture center (CCM).
When mentioning \emph{El Raval}, tourists publications mention its museums and other nearby districts. On the contrary, locals publications talk about its cultural activity, its promenade or its drug presence problem.

\subsection{Beyond Districts}
The trained Word2Vec models provide information that can be used beyond a district analysis. They can infer the words that Instagram users relate to Barcelona and any other word in the training vocabulary. For the following queries, the translation of the English query word to the corresponding language has been used, and the translation of the local languages results to English are shown.\\

\begin{multicols}{2}
\scriptsize
\textit{Beach}\\
\textcolor{red}{summer, seaside, sand, sunset, sunny, whotel, seaview}\\
{\color{green!70!blue}seaview, passeigmaritim, mediterráneo, novaicaria}\\
\textcolor{blue}{bogatell, novaicaria, mediterrani, lamarbella, voley}\\

\textit{Food}\\
\textcolor{red}{tapas, fastfood, breakfast, sangria, seafood, blackrice}\\
{\color{green!70!blue}sushi, ham, soup, chicken, hamburguer, fruit}\\
\textcolor{blue}{cannelloni, omelette, bread, soup, fruit, fries}\\

\textit{Tourism}\\
\textcolor{red}{landscape, architecture, bluesky, spain, aroundtheworld}\\
{\color{green!70!blue}catalanfood, adventure, crossing, eatanddrink, fair}\\
\textcolor{blue}{route, beatifullplaces, mycity, heritage, walking}\\

\textit{Neighborhood}\\
\textcolor{red}{quaint, hidden, restaurants, lively, locals, gracia, corners}\\
{\color{green!70!blue}gotico, turists, citizens, park, streets, shops, people}\\
\textcolor{blue}{ribera, citizen, ravalnotforsale, citizenfight, street, walking}\\
\end{multicols}

This experiments also show clear differences between the models trained with the different languages. For instance, when mentioning \emph{Food} English speakers write along Spanish most characteristic dishes, while locals write about more daily meals. When mentioning \emph{Neighborhood}, tourists talk about its restaurants or appearance, while locals talk more about its people.

\section{Visual Analysis}
An image worths a thousand words. Word2Vec allows us to find the words that authors relate neighborhoods when using different languages. 
That is possible because Word2Vec learns word embeddings to a vectorial space where semantic similar words (words appearing in similar contexts), are mapped nearby. Img2NeighCtx (Image to Neighborhood Context) is a Convolutional Neural Network that, learning from images and associated captions, allows us to find the images that authors relate to the different neighborhoods when using different languages. 

\begin{figure*}[h]
	\centering
  \includegraphics[width=0.9\linewidth]{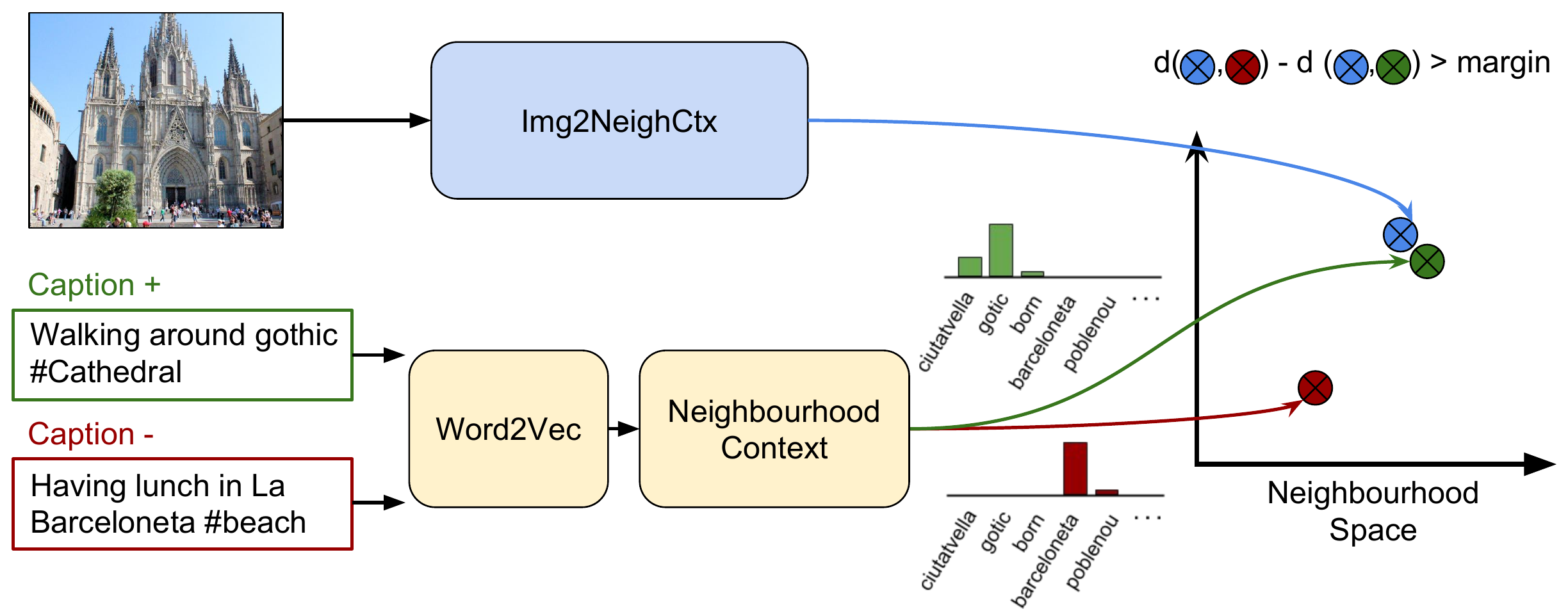}
   \caption{Training procedure of Img2NeighCtx. The CNN is trained to maximize the difference of distances between the image and the positive caption and the image and the negative caption in the \emph{Neighborhood Space} space until a certain margin.}
   \label{fig:Img2NeighCtx_training}
\end{figure*}

\subsection{Img2NeighCtx} \label{sec:51}
Word2Vec allows us to compute a similarity between two words. To compute a vector encoding the similarities of a caption with each of the $82$ Barcelona's districts and neighborhoods, we sum up the cosine similarities of all the caption words with each neighborhood name in the Word2Vec space and L2 normalize the vector. We call the resulting vector \emph{Neighborhood Context} ($NC$). Let $W$ be the Word2Vec representations of all the words in a caption $c$ and $N = \{n_j\}_{j=1:J}$ be all the neighborhoods names Word2Vec representations ($J=82$). The neighborhood context of each word $w$ in the caption is represented by
\begin{equation}
NC(w) = \left(\frac{\langle w, n_1 \rangle}{||w|| \cdot ||n_1||}, \frac{\langle w, n_2 \rangle}{||w|| \cdot ||n_2||}, \dots , \frac{\langle w, n_J \rangle}{||w|| \cdot ||n_J||} \right) 
\end{equation}
We eventually compute the \emph{Neighborhood Context} of the caption $c$ as:
\begin{equation}
NC(c) = \sum_{w \in W} NC(w)
\end{equation}
which is L2 normalized.


Img2NeighCtx is a GoogleNet based CNN that learns to infer $NC$ from images.
The last classification layer is replaced by a fully connected layer with $82$ outputs, which is the dimensionality of the \emph{Neighborhood Space}, and uses a ranking loss to learn to embed images with similar captions \emph{Neighborhood Contexts} nearby. Img2NeighCtx receives three inputs: the image ($i$), its caption embedding ($NC^{+}$), and a negative caption embedding ($NC^{-}$). The negative caption embedding is selected randomly from the 
50\% most distant batch caption embeddings. We define the loss by
\begin{equation}
L(i,NC^{+},NC^{-}) = \tfrac{1}{2} max\left(0,m - \Phi ^{T}_{i}NC^{+} + \Phi ^{T}_{i}NC^{-}\right)
\end{equation}
where $m$ is the margin and $\Phi$ is the function that embeds the image into the \emph{Neighborhood Space}.

Img2NeighCtx is trained to minimize this loss, which maximizes the difference between the distances of the image with the positive and negative captions upon a certain margin. The training pipeline of Img2NeighCtx is shown in Figure \ref{fig:Img2NeighCtx_training}.
The weights are initialized with an ImageNet \cite{Deng} pretrained network, and is trained using Stochastic Gradient Descent with a learning rate of $0.001$, 
a momentum of $0.9$ and a weight decay of $2e-4$. The margin is set empirically to $0.4$. We trained an Img2NeighCtx model for each one of the languages using a batch size of $120$ and the three models converged around $100,000$ iterations.

To ensure that Img2NeighCtx learns to recognize generic visual features instead of overfitting to the training data, we randomly split each language dataset in three subsets: $80\%$ training set, $5\%$ validation set and $15\%$ retrieval set. The validation set is used to monitor overfitting when training the model. The retrieval set is not used to train, but to test the model in the following experiments.
This configurations ensures that the trained models can generalize beyond the data used in this work.

\subsection{Images Associated to Districts} \label{sec:52}
\begin{figure}[h]
	\centering
  \includegraphics[width=0.98\linewidth]{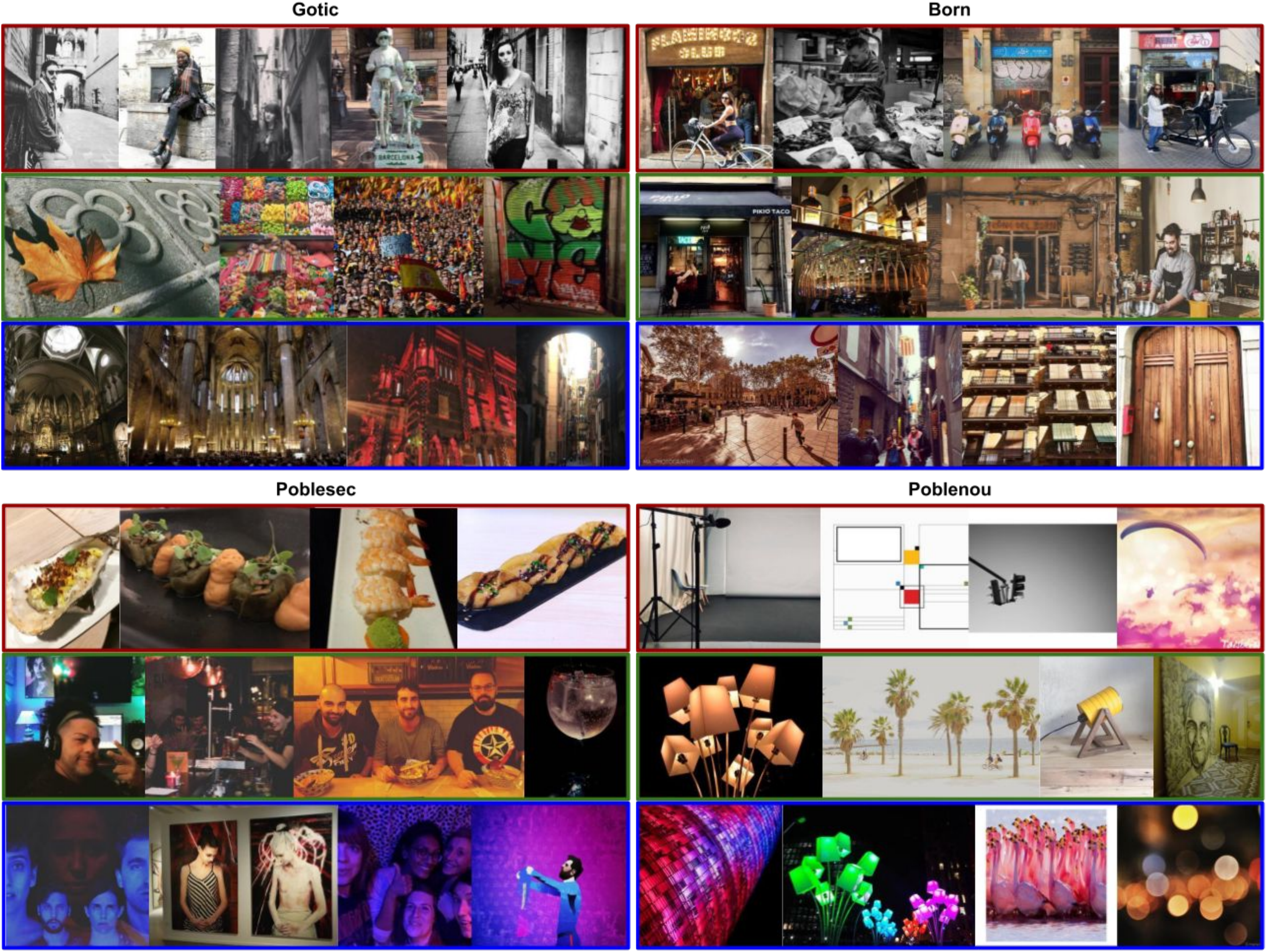}
   \caption{{Img2NeighCtx} image by neighborhood retrieval results for different neighborhoods in each of the languages.}
   \label{fig:txt2img_neigh_all}
\end{figure}

Once Img2NeighCtx has been trained to embed images in the \emph{Neighborhood Space}, it can be used in a straightforward manner in 
an image by neighbourhood retrieval task.
The CNN has learned from the images and the associated captions to extract visual features useful to relate images to the different neighborhoods. 
Using as a query a neighborhood represented as a one hot vector in the \emph{Neighborhood Space}, we can infer the kind of images that Instagram users writing in English, Spanish or Catalan relate to that neighborhood. To do that we retrieve the nearest images in the \emph{Neighborhood Space}. Figures \ref{fig:txt2img_barceloneta} and \ref{fig:txt2img_neigh_all} show the first retrieved images for some of the neighborhoods. Images in the top row (red) correspond to the English trained model, in the second (green) to the Spanish one, and in the third (blue) to the Catalan one.
When posting about \emph{La Barceloneta} (Fig. \ref{fig:txt2img_barceloneta}), tourists tend to post photos of the drinks they have at the beach, while locals tend to post photos of themselves posing. When talking about \emph{El Born} (\emph{Sant Pere}) (Fig. \ref{fig:txt2img_neigh_all}), tourist tend to post photos of bikes, since there are many tourist oriented stores offering bike renting services there, while locals tend to post photos of its bars and streets. When posting about \emph{El Poblesec} \ref{fig:txt2img_neigh_all}, tourist tend to post photos of the food they have in its popularity increasing restaurants, while locals tend to post photos of themselves, its bars or its art galleries. When posting about \emph{El Poblenou} \ref{fig:txt2img_neigh_all}, the kind of images  people post using the three languages are similar and related to design and art. This is because \emph{El Poblenou} neighbourhood has been promoted as a technology and design hub in Barcelona, following the 22@ plan. This plan has attracted many foreign workers to live in the area. Therefore, and in contrast to other neighborhoods, the majority of English publications related to \emph{El Poblenou} are not from tourists but from people that have settled here, and appear to have the same interests in \emph{El Poblenou} as the Catalan and Spanish speakers. 

\subsection{Beyond Districts: Img2Word2Vec}
Img2NeighCtx is very useful to retrieve images associated to each neighborhood in each one of the languages. In a similar way we trained Img2NeighCtx to predict \emph{Neighborhood Contexts} from images, we can train a net to directly embed images in the Word2Vec space. We call that net Img2Word2Vec.

First, the embeddings of all the captions in the Word2Vec space are computed as the mean of its word embeddings and L2 normalized. Img2Word2Vec has the same structure as Img2NeighCtx, but the last fully connected layer has $300$ outputs, which is the dimensionality of the Word2Vec space. It uses a ranking loss to learn to embed semantically similar images nearby. Img2Word2Vec receives 3 inputs: the image, its caption Word2Vec embedding, and a negative caption Word2Vec embedding. The negative caption embedding is selected randomly from the 
other batch captions. The training pipeline is similar to the Img2NeighCtx one (Figure \ref{fig:Img2NeighCtx_training}) but leaving out the Neighborhood Context computation and applying the ranking loss directly to captions Word2Vec embeddings.
We trained one Img2Word2Vec model for each language. The splits and the training parameters used were the same as for Img2NeighCtx. All the models converged around $150,000$ iterations.

\begin{figure}[h]
	\centering
  \includegraphics[width=0.98\linewidth]{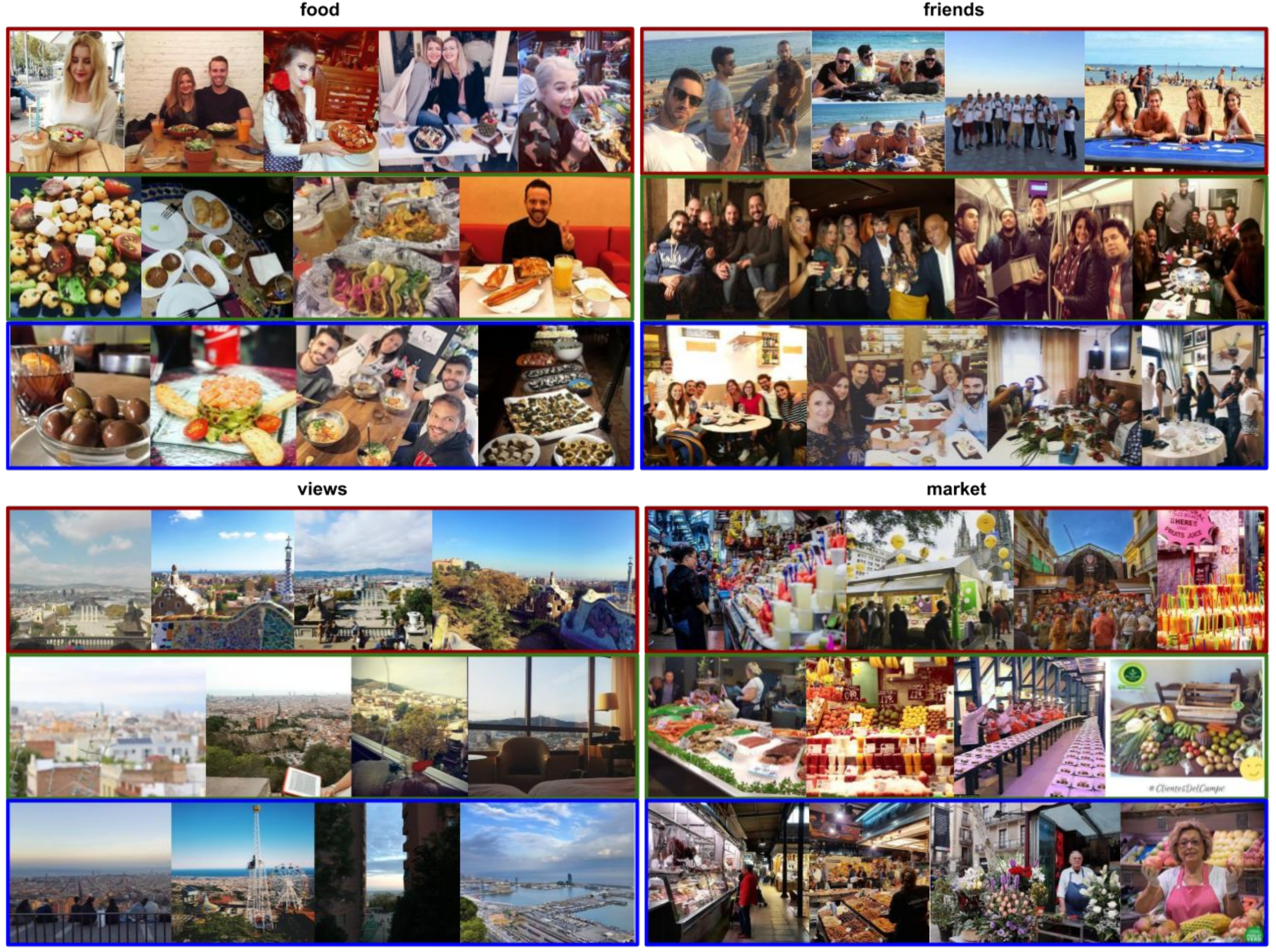}
   \caption{{Img2Word2Vec} image by text retrieval results for different queries in each of the languages.}
   \label{fig:txt2img_all}
\end{figure}

The trained Img2Word2Vec models can be used to relate text and images beyond districts and neighborhoods names, retrieving images related to any text concept present in the vocabulary. Figure \ref{fig:txt2img_all} shows retrieval results for different query words. When using the word \emph{food}, tourist tend to post photos of themselves in front of "healthy" and well presented dishes or seafood. As a contrast, locals tend to post photos where only the food appears, and it tends to be international and more diverse. 
For \emph{friends} tourist tend to post photos of a group of friends in the beach, while locals tend to appear around a table, though they are more diverse. Associated with the word \emph{views}, tourists post photos of Barcelona's views taken from popular places (Montjuic and Park Güell). As a contrast, locals photos are more diverse and include photos taken from houses and of other Barcelona areas, such as the port. When using the word \emph{market}, tourist photos are mainly from Mercat de la Boqueria, an old market in Barcelona's old town that has turned into a very touristic place. Meanwhile, locals photos are more divers and include markets where people do their daily shopping.
In general, English speakers images are much less variant than local languages speakers images, and more concentrated in popular spots. That proves that the assumption that English speakers images correspond mainly to tourists is true, and also that tourism is strongly concentrated in certain Barcelona areas, as \cite{Garcia-Palomares2015} also concluded.

\section{Conclusions}
Extensive experiments have demonstrated that Instagram data can be used to learn relations between words, images and neighborhoods that allow us to do a per neighborhood analysis of a city. Results have shown that the assumption that English publications represent tourists activity and local languages publications correspond to locals activity is true. Both the textual and the visual analysis have demonstrated to reflect the actual tourists and locals behavior in Barcelona.

The retrieval results for both Img2NeighCtx and Img2Word2Vec nets have been obtained in blind and image only test sets, which proves that similar results can be obtained with external images. Moreover, Img2Word2Vec can be used to obtain results for any term in the vocabulary. In this work the \emph{InstaBarcelona} dataset has been used. However, models can be scaled to larger datasets, since both Word2Vec and CNNs scale well with big data. The experiments can also be extended straightforward to other cities or subjects. 
The code used in the project is available on \url{https://github.com/gombru/insbcn}.

\section*{Acknowledgments}
This work was supported by the Doctorats Industrials program from the Generalitat de Catalunya, the Spanish project TIN2017-89779-P, the H2020 Marie Skłodowska-Curie actions of the European Union, grant agreement No 712949 (TECNIOspring PLUS), and the Agency for Business Competitiveness of the Government of Catalonia (ACCIO).
%
%
%
%
\bibliographystyle{splncs04}
\bibliography{mybib}

\begin{thebibliography}{10}
\providecommand{\url}[1]{\texttt{#1}}
\providecommand{\urlprefix}{URL }
\providecommand{\doi}[1]{https://doi.org/#1}

\bibitem{bcn_hotel_beds}
{Ayuntament de Barcelona. Barcelona stadistics. Observatory districts.}
  (2017),
  \url{http://www.bcn.cat/estadistica/angles/documents/districtes/index.htm}

\bibitem{Blei2003}
Blei, D.M., Ng, A.Y., Jordan, M.I.: {Latent Dirichlet Allocation}. J. Mach.
  Learn. Res.  (2003)

\bibitem{Boy}
Boy, J.D., Uitermark, J.: {Reassembling the city through Instagram}. Trans.
  Inst. Br. Geogr.  (2017)

\bibitem{Garcia-Palomares2015}
Garc{\'{i}}a-Palomares, J.C., Guti{\'{e}}rrez, J., M{\'{i}}nguez, C.:
  {Identification of tourist hot spots based on social networks: A comparative
  analysis of European metropolises using photo-sharing services and GIS}.
  Appl. Geogr.  (2015)

\bibitem{Gomez2017}
Gomez, L., Patel, Y., Rusi{\~{n}}ol, M., Karatzas, D., Jawahar, C.V.:
  {Self-supervised learning of visual features through embedding images into
  text topic spaces}. CVPR  (2017)

\bibitem{DianeLarlus2017}
Gordo, A., Larlus, D.: {Beyond Instance-Level Image Retrieval: Leveraging
  Captions to Learn a Global Visual Representation for Semantic Retrieval}.
  CVPR  (2017)

\bibitem{Chang}
He, Y., Yang, X., Zhang, X.: {Instagram Post Data Analysis}. arXiv  (2015)

\bibitem{Engineering2017}
Instagram: {Instagram's Neighborhood Flavors – Instagram Engineering}. Medium
   (2017)

\bibitem{Deng}
{Jia Deng}, {Wei Dong}, Socher, R., {Li-Jia Li}, {Kai Li}, {Li Fei-Fei}:
  {ImageNet: A large-scale hierarchical image database}. CVPR  (2009)

\bibitem{Kuo}
Kuo, Y.H., Hung, C.Y., Hsieh, L.C., Hsu, W., Chen, Y.Y., Chen, B.C., Lee, W.Y.,
  Wu, C.C., Lin, C.H., Hou, Y.L., Cheng, W.F., Tsai, Y.C.: {Discovering the
  City by Mining Diverse and Multimodal Data Streams}. ACM Int. Conf. Multimed.
   (2014)

\bibitem{Mikolov2013}
Mikolov, T., Corrado, G., Chen, K., Dean, J.: {Efficient Estimation of Word
  Representations in Vector Space}. ICLR  (2013)

\bibitem{Frome2013}
Norouzi, M., Mikolov, T., Bengio, S., Singer, Y., Shlens, J., Frome, A.,
  Corrado, G.S., Dean, J.: {Zero-Shot Learning by Convex Combination of
  Semantic Embeddings}. NIPS  (2013)

\bibitem{Pennington}
Pennington, J., Socher, R., Manning, C.: {Glove: Global Vectors for Word
  Representation}. EMNLP  (2014)

\bibitem{Salvador}
Salvador, A., Hynes, N., Aytar, Y., Marin, J., Ofli, F., Weber, I., Torralba,
  A.: {Learning Cross-Modal Embeddings for Cooking Recipes and Food Images}.
  CVPR  (2017)

\bibitem{Singh2017}
Singh, V.K., Hegde, S., Atrey, A.: {Towards measuring fine-grained diversity
  using social media photographs}. ICWSM  (2017)

\end{thebibliography}
\end{document}